# Research Highlights

- CLPSO-ML modeling framework is proposed for Influenza Prediction.

- Three multi-step-ahead prediction modeling strategies were compared and justified.

- The MIMO strategy achieves the best prediction performance with for longer horizon.

# Comprehensive Learning Particle Swarm Optimization Enabled Modeling Framework for Multi-step-ahead Influenza Prediction

Siyue Yang, Yukun Bao[*]

Center for Modern Information Management,

School of Management, Huazhong University of Science and Technology, Wuhan 430074, P.R. China

## Abstract

Epidemics of influenza are major public health concerns. Since influenza prediction always relies on the weekly clinical or laboratory surveillance data, typically the weekly Influenza-like illness (ILI) rate series, accurate multi-step-ahead influenza predictions using ILI series is of great importance, especially, to the potential coming influenza outbreaks. This study proposes Comprehensive Learning Particle Swarm Optimization based Machine Learning (CLPSO-ML) framework incorporating support vector regression (SVR) and multilayer perceptron (MLP), two well-established machine learning models, for multi-step-ahead influenza prediction. A comprehensive examination and comparison of the performance and potential of three commonly used multi-step-ahead prediction modeling strategies, including iterated strategy, direct strategy and multiple-input multiple-output (MIMO) strategy, was conducted using the weekly Influenza-like illness (ILI) rate series from both the Southern and Northern China. The results show that: 1) The MIMO strategy achieves the best multi-step-ahead prediction, and is potentially more adaptive for longer horizon; 2) The iterated strategy demonstrates special potentials for deriving the least time difference between the occurrence of the predicted peak value and the true peak value of an influenza outbreak; 3) For ILI in the Northern China, SVR model implemented with MIMO strategy performs best, and SVR with iterated strategy also shows remarkable performance especially for accurate prediction during outbreak periods; while for ILI in the Southern China, both SVR and MLP models with MIMO strategy have competitive prediction performance.

---

[*] Corresponding author: Tel: +86-27-87558579; Fax: +86-27-87556437.

Email: yukunbao@hust.edu.cn or y.bao@ieee.org

# 1. Introduction

Seasonal influenza is an acute respiratory infection caused by influenza viruses, which circulates worldwide and remains a serious public health problem. According to World Health Organization, influenza epidemics are estimated to cause about 3 to 5 million severe illness cases and about 290,000 to 650,000 respiratory deaths each year. During outbreak seasons, hospitals and public disease control departments are under huge pressure to take countermeasures such as medical resources allocation and vaccination campaigns. The decisions and plans above are made mainly on the basis of clinical and laboratory surveillance data issued by local and national centers of disease control, typically weekly influenza-like illness (ILI) rate series. Nevertheless, the released surveillance data intrinsically describes influenza occurred in the past, limiting their utility for public health decision making[1]. Considering the annually distinct timing and intensity of influenza epidemics, accurate prediction of influenza and its outbreak with longer time horizon in advance could provide reliable epidemic signals for public health response. As mentioned in[2], improving influenza prediction continues to be a central priority of global health preparedness efforts.

A variety of influenza prediction approaches have been summarized and evaluated in literature reviews[3–5] and in the documentary works of the Centers for Disease Control and Prevention (CDC) Challenges and FluSight projects[1,2]. Time series models, which predict the trend of influenza in the future through analyzing the potential temporal relationships and outbreak patterns of historical data, remain dominant in extant researches. Traditional statistical time series models widely applied to forecast the influenza, include Autoregressive integrated moving average (ARIMA)[6–11], Generalized linear model (GLM)[7,12,13], Least absolute shrinkage and selection operator (LASSO)[7,8,14] and other types of regression models[15–17]. Due to the boom of data-driven technology, machine learning models have been employed in time series forecasting, demonstrating the superiority in modeling complex non-linear relationships between target and dependent variables[18]. As for influenza prediction, Support vector regression (SVR)[11,13,19,20], Random forest (RF) [11,13,21], Gradient boosting (GB) [11,13] and deep learning models such as Long short-term memory (LSTM)[9,13,19,22–24] have been exploited in influenza prediction. In addition, ensemble approaches gathering multiple models become popular in recent influenza prediction research, including Bayesian model averaging (BMA)[7,8] , stacking method[11,25,26] and other approaches[14,27]. Combining multiple models has been proved to improve the comprehensive forecasting performance compared to single models in abovementioned studies.

Early researches on influenza prediction mainly focused on one-step-ahead prediction, namely using all or some of the observations to estimate a variable of interest for the time-step immediately following the latest observation[28]. While in practice, influenza outbreak seasons always last more than one month, and thus, the longer the time span before influenza outbreak is predicted, the more sufficiently the preparation and arrangement for hospital resources can be undertaken. Therefore, multi-step-ahead prediction, namely predicting a sequence of values ahead the latest observation[29], is necessarily required to support public health decision-making against influenza epidemics. In recent years, multi-step-ahead prediction has been taken into consideration in influenza prediction[9–11,16,23,25,26]. However, specific attention on different multi-step-ahead modelling strategies was rarely paid in prior researches when implementing influenza prediction models, and the dominating strategy by default was iterative only. As Taieb et al.[30] concluded, there should be at least three major modelling strategies for the multi-step-ahead prediction. In the iterated strategy (also called recursive strategy), multi-step-ahead prediction is conducted by training a one-step-ahead forecasting model and subsequently iterating the one-step-ahead forecast $h$ times to achieve $h$-step-ahead prediction. In the direct strategy, $h$ single-output forecasting models are separately trained for each step using the same observations as input variables, which leads to huge computational costs. As per the multiple input multiple output (MIMO) strategy, one multi-output model is trained to forecast the whole horizon in one shot, with predicted values in form of a vector rather than a scalar quantity. The MIMO strategy is available for machine learning models with intrinsic multiple output structure and unsuitable for most traditional linear regression models. As is studied in energy area[31] and with simulated data[28,30], different strategies could significantly influence the multi-step prediction performance with the same data sets, especially in the cases of machine learning models. On the other hand, most existing multi-step-ahead influenza prediction models are evaluated merely using averaged statistical metrics such as mean square error (MSE), root mean squared error (RMSE) and maximum absolute percent error (MAPE). In practice, accurate prediction aiming at the timing and severity of influenza outbreak peaks is highly concerned and quite important for decision-making. Thus, the prediction performance during influenza outbreak periods ought to be measured with close attention in order to fully evaluate the different multi-step-ahead prediction models at hands.

In order to assure the high quality of multi-step-ahead prediction models with different strategies, model tuning process including feature selection and parameter optimization should be involved in the modeling framework. However, in extant researches on influenza prediction with machine learning (ML) models, the hyperparameters are mostly predefined rather than being optimized. As for feature selection, input features are mainly determined either following experience of experts or with some filter methods. For instance, Cheng et al.[11] identified the input features according to suggestions of experts in Taiwan CDC; Darwish et al.[13] proposed three types of feature space integrating time lags and first-order differences of weekly ILI rate series. An enumeration in small range was taken to find the best feature space. Liu et al.[22] calculated Pearson parametric correlation matrix between candidate features and ILI rate, and selected the features having significant correlation with the predicted variable. While in multi-step-ahead influenza prediction, even if the same ML model and candidate input features are applied to predict the same horizon of ILI rates, the training and predicting procedures differ among multi-step-ahead strategies, and thereby the optimum features and hyperparameters should be identified respectively to improve forecasting accuracy of each model implemented with each strategy. Such tasks could not be accomplished effectively and efficiently by the predefined hyperparameters or the filter methods selecting features merely.

In recent years, hybrid modeling frameworks combining forecasting methods with optimization techniques for model tuning including feature selection and hyperparameters tuning have been widely exploited and proved to improve forecasting accuracy significantly and efficiently in many domains such as energy, economics, engineering as well as disease detection[32–39]. For example, Kumar and Susan[32] built a hybrid modeling framework using fuzzy time series (FTS) forecasting and particle swarm optimization (PSO) algorithm to predict COVID-19 pandemic, where

PSO was applied to optimize the values of three hyperparameters for FTS. Two specific frameworks "nested FTS-PSO" and "exhaustive search FTS-PSO" were proposed and examined on the dataset of coronavirus confirmed cases from 10 countries, and the exhaustive search FTS-PSO performed best. Altan et al.[33] developed a hybrid modeling framework for wind speed forecasting, combining long short-term memory (LSTM) network and decomposition methods with grey wolf optimizer (GWO), a swarm intelligence-based meta-heuristic algorithm to optimize the intrinsic mode function estimated outputs. The superiority of the proposed forecasting framework was validated on the data from five wind farms in the Marmara region, Turkey. Karasu et al.[34] proposed a forecasting framework for crude oil price based on support vector regression (SVR) with a wrapper method of feature selection using multi-objective PSO, considering both MAPE and Theil's U values as evaluation metrics. Zhang and Lim[35] constructed an ensemble transfer learning framework for optic disc segmentation with an improved PSO-based hyperparameter identification method to optimize the learning parameters in Mask R-CNN model. Evaluated on the Messidor and Drions image datasets, the proposed ensemble framework proved to significantly outperform other ensemble models integrated with original and advanced PSO variants. From previous studies, it can be found that PSO and its variants are successfully and frequently applied in plenty of existing model tuning tasks. However, the majority of extant solutions tended to treat parameter optimization and feature selection as two separate subtasks, and always focused on refining one of them in their proposed hybrid framework. In this study, to efficiently achieve the model tuning tasks in multi-step-ahead influenza prediction, we develop a unified adaptive modeling framework which is able to deal with parameter optimization and feature selection simultaneously in one optimization process, taking into account the mutual influence between choice of feature subsets and appropriate hyperparameters under different multi-step strategies. Since the unified process leads to increased complexity of the optimization problem and the traditional binary PSO was pointed out to suffer from trapping into local optimal solutions in feature selection[40], we adopt Comprehensive Learning Particle Swarm Optimization (CLPSO) in binary form as optimizing algorithm. As a powerful variant of PSO, CLPSO preserves the diversity of swarm, discourages premature convergence and thus improves the global search ability of PSO. Meanwhile, validated by [41] that initially proposed this variant, CLPSO demonstrates significant superiority in solving multimodal problems, which indicates its fitness for complex optimization problems.

In order to make original CLPSO fit in our modeling framework, we design a binary encoding approach for particles to represent the selected features and hyperparameters simultaneously, which is critical to realize the unified model tuning process.

In a nutshell, the main aim of this study is to explore the performance and potential of machine learning models implemented with different multi-step-ahead modelling strategies in multi-step-ahead influenza prediction, which can provide a comprehensive and reliable reference for practitioners and researchers to applying machine learning models for multi-step-ahead influenza prediction in the future, and thus could further support the decision making of public disease control departments, hospitals, and pharmaceutical companies. The contributions of this work are summarized as follows:

- This study firstly takes into account the implementation and comparison of different multi-step-ahead modelling strategies on multi-step-ahead influenza prediction with machine learning models. Although some extant researches have predicted influenza in a multi-step-ahead way using machine learning models with multi-output structures, few, if any, studies have paid attention to examine the diverse performance of a specific machine learning model with different multi-step-ahead strategies.
- To efficiently tune the models involved in the multi-step-ahead influenza prediction and make it perform as well as possible, this study proposes a Comprehensive Learning Particle Swarm Optimization based Machine Learning (CLPSO-ML) framework as a unified adaptive model tuning process of both feature selection and parameters optimization. Equipped with the CLPSO-ML model tuning approach, multi-step-ahead influenza prediction can be improved well.
- More comprehensive evaluations are conducted for multi-step-ahead influenza prediction results, considering the outbreak timing and magnitude accuracy of influenza. And thus the decision-making toward the outbreak of influenza can be well supported.

The rest of this paper are organized as follows. Section 2 provides a brief review of the literature on influenza prediction. In Section 3, multi-step-ahead prediction strategies, the selected two well-established machine learning models and the proposed CLPSO-ML modeling framework are illustrated in details. Section 4 describes the experiment design including data source, data preprocessing, experimental procedure, evaluation of model performance and other implementation details. Experimental results are discussed in Section 5. Finally, conclusions are drawn in Section 6.

## 2. Literature review

This section briefly reviews the literature on influenza prediction using statistical approaches including both traditional regression models and machine learning models. Table 1 provides an overview of those studies.

In terms of prediction target (predicted variable), Influenza-Like Illness (ILI) rate series is regarded as the proxy of confirmed influenza cases and serves as the prediction target in the overwhelming majority of the extant researches, which is generally monitored and issued by local or national CDC in different countries. ILI refers to the case that a patient has fever (temperature of 100°F [37.8°C] or greater) and a cough and/or a sore throat without a known cause other than influenza. Among those studies, ILI of multiple spatial scales in the United States has been mostly considered, including level of nation[6,10,16,17,25,26], state[8,22,27] and region defined by Health and Human Services (HHS)[9,10,25,26]. ILI in China was also predicted in many cases, mainly at the scale of city and province such as Guangzhou[24], Shenzhen[21], Hongkong[7], Liaoning[20] and Taiwan[11]. However, larger regional scale of ILI in China is in lack of exploration in previous researches. In this study, ILI rate series from the Southern and Northern China issued by Chinese National Influenza Center are used as data sets.

As per the prediction horizon, about half of the reviewed literature focused on nowcasting and one-step-ahead prediction. The rest of studies considered longer prediction horizon, namely multi-step-ahead prediction. It is mentioned in [30] that different multi-step-ahead modelling strategies can be suitable for different length of prediction horizon. For instance, it was discovered in [42] that iterated strategy beats the other two strategies for long-term flood forecasting with neural network model. Ghysels ezxt al. [43] compared the direct strategy, the iterated strategy and a mixed-data sampling strategy for multi-step-ahead volatility forecasting, and found that iterated strategy performs best at short horizons while the mixed-data sampling strategy has an edge for longer horizons. However, previous researches on influenza prediction conducted multi-step ahead prediction merely with one predefined strategy (iterated strategy mostly) ignoring the comparison among different multi-step-ahead modelling strategies. And thus this study conducted an extensive experiment to evaluate the overall performance of multi-step-ahead modelling strategies across different time horizons ranging from two-step-ahead to ten-step-ahead for influenza prediction.

As per evaluation metrics, almost all previous researches on multi-step-ahead influenza prediction did not have a close examination on the prediction performance during influenza outbreak periods but the overall time window before and after the outbreak only. According to extant researches, averaged statistical metrics including Mean Absolute Error (MAE), Root Mean Squared Error (RMSE), Root Mean Squared Percent Error (RMSPE), Mean Absolute Percentage Error (MAPE) and Mean Squared Error (MSE) are commonly applied for evaluating prediction performances. However, metrics demonstrates the prediction performance during the influenza outbreak should be taken into account as well. Therefore, Peak Magnitude Error (PME) and Peak Week Error (PWE, also called Week Difference)[7,17], which directly calculate the time difference and magnitude difference between the true peak value and the predicted peak value, were taken as the metrics in this present study. According to Chinese National Influenza Center, influenza viruses are extremely active in winter due to the weather condition, which provides a rough definition of outbreak seasons. Thus, influenza outbreak seasons are defined as the periods between the $45^{th}$ week in the current year and the $8^{th}$ week in the next year. Based on the definition of influenza outbreak season for ILI in China, in this study, PWE is adopted to measure the peak time difference, while Mean Absolute Error (MAE) is additionally calculated to evaluate the prediction accuracy of the whole outbreak period.

In summary, different from the existing literature, this study contributes to the influenza prediction literature by concentrating on the application and comparison of different modelling strategies for multi-step-ahead influenza prediction based on machine learning models with ILI data in the Southern and Northern China, which has not yet been fully explored in previous researches. Furthermore, this study considers analyzing the fitness of different multi-step strategies under different length of prediction horizon, and additionally takes the influenza outbreak related accuracy metrics into account for experimental justification, both of which were ignored in extant multi-step-ahead influenza prediction researches.

Table 1 Overview on influenza prediction literature

| Authors | Modeling Technique(s) [a] | Prediction target | Time scale | Prediction Horizon | Evaluation Metrics [b] | Model type [c] |
|---|---|---|---|---|---|---|
| Dugas et al.[12] | GLM, GARMA | Influenza cases | Weekly | 1 | GFD, FC | TS |
| Preis and Moat[6] | ARIMA | ILI | Weekly | nowcasting | MAE | TS |
| Volkova et al.[19] | Adaboost, SVR, LSTM | ILI | Weekly | 2 | PC, RMSE, RMSPE, MAPE | ML |
| Wu et al.[21] | RF | ILI | Weekly | 1 | MAPE, MSE | ML |
| Xu et al.[7] | GLM, LASSO, ARIMA, DL with FNN, model ensemble using BMA | ILI | Weekly | 2 | MAPE, RMSE, MAE, WD | TS & ML |
| Yang et al.[16] | MLR | ILI | Weekly | 3 | RMSE, MAE, RMSPE, MAPE, correlation | TS |
| Brooks et al.[25] | Delta density model with stacking-based adaptively weighted ensemble methods | ILI; | weekly | 4 | Unibin log score, Multibin log score, AE | TS |
| Ertem et al.[17] | EBF, MLR | ILI | weekly | 1 | RMSE, PWE, PME | TS & ML |
| Liang et al.[20] | SVR | Influenza cases | Monthly | 1 | RMSE, RMSPE, MAPE | ML |
| Liu et al.[22] | LSTM | ILI | Weekly | 1 | RMSE | ML |
| Soliman et al.[8] | DL with FNN, Beta regression, ARIMA, LASSO, MARS; model ensemble using BMA | ILI | Weekly | 2 | RMSE, MAPE, MAE, CRPS | TS & ML |
| Venna et al.[9] | LSTM-based multi-stage model, ARIMA, EAKF | ILI, Google Flu Trend | Weekly | 5 | MAPE, RMSE, RMSPE | TS & ML |
| Lu et al.[27] | ARGO, ARGONet | ILI | Weekly | nowcasting | MAPE, RMSE, PC | TS & ML |
| Kandula et al.[10] | ARIMA, ARIMA-STL, AR-NN | ILI | Weekly | 4 | Log score, MAPE | TS & ML |
| Reich et al.[26] | A stacking ensemble model combined 21 component models | ILI | Weekly | 4 | Exponentiated average log scores, RMSE, Average bias | TS & ML |
| Zhu et al.[24] | attention-based multi-channel LSTM | ILI | Weekly | 1 | MAPE | ML |
| Darwish et al.[13] | GLM, SVR, GB, RF, LSTM | ILI | Weekly | 1 | MAPE, RMSE | ML |

| Cheng et al.[11] | ARIMA, RF, SVR, GB; stacking ensemble approach | ILI | Weekly | 3 | PC, RMSE, MAPE, Hit rate | TS & ML |

[a] GLM: Generalized Linear Models, GARMA: Generalized linear AutoRegressive Moving Average, SVR: Support Vector Regression, LSTM: Long Short Term Memory, RF: Random Forest, LASSO: Least Absolute Shrinkage and Selection Operator, ARIMA: AutoRegressive Integrated Moving Average, DL with FNN: Deep Learning with Feedforward Neural Networks, BMA: Bayesian Model Averaging, MLR: Multivariate Linear Regression, EBF: Empirical Bayes Framework, BR: Beta Regression, MARS: non-parametric Multivariate Adaptive Regression Splines, EAKF: Ensembled Adjustment Kalman Filter, ARGO: AutoRegression with General Online information, ARGONet: Network-based ensemble approach with AutoRegression with General Online information, ARIMA-STL: ARIMA with Seasonal and Trend Decomposition, AR-NN: feed-forward AutoRegressive artificial Neural Network, GB: Gradient Boosting.
[b] GFD: Global Forecast Deviance, FC: Forecast Confidence, MAE: Mean Absolute Error, PC: Pearson Correlation, RMSE: Root Mean Squared Error, RMSPE: Root Mean Squared Percent Error, MAPE: Mean Absolute Percentage Error, MSE: Mean Squared Error, WD: Week Difference, AE: Absolute Error, PWE: Peak Week Error, PME: Peak Magnitude Error, CRPS: Continuous Ranked Probability Score.
[c] TS: Traditional Statistical model, ML: Machine Learning model.

## 3. Methodologies

Notations are illustrated as follows in the context of influenza prediction at the beginning of this section. Given the weekly ILI rate series $\{I_1, I_2, \ldots, I_N\}$, the current and previous observations are denoted as $x = [I_t, I_{t-1}, \ldots, I_{t-d+1}] \in \mathbb{R}^d$, which represents that the input features are the ILI rates from the week $t - d + 1$ to the current week $t$, with the embedded dimension $d$. Whereas the future observation is denoted as $y = [I_{t+1}, I_{t+2}, \ldots, I_{t+H}] \in \mathbb{R}^H$, which represents that the predicted values are the future ILI rates from the week $t + 1$ to the week $t + H$, with $H$ as the length of time horizon of prediction. In a word, multi-step-ahead influenza prediction aims to find the functional relationship between the current and previous ILI rates, namely $x$, and the future ILI rates, namely $y$.

### 3.1 Multi-step-ahead modelling strategies

In this study, three modelling strategies are applied for multi-step-ahead influenza prediction, namely iterated strategy, direct strategy, and MIMO strategy. As there are various variations of the strategies with specific situations and models in extant literature, it is impossible and unnecessary to enumerate each variation. Thus, this section describes the standard version of each strategy.

### 3.1.1 Iterated strategy

In the iterated strategy, a single model $f$ is trained by minimizing the error for one-step-ahead prediction as follows:

$$I_{t+1} = f(I_t, \dots, I_{t-d+1}) + w \qquad (1)$$

Where $d$ denotes embedded dimension, and $w$ denotes the additive noise.

For $H$-step-ahead prediction, the one-step-ahead prediction is achieved using the trained model, and then the predicted value is fed into the trained one-step-ahead model when forecasting the second step after current time t. The above process is iterated to predict subsequent steps until reaching the $H$ step. The process above is formulated as follows:

$$\hat{I}_{t+h} = \begin{cases} \hat{f}(I_t, I_{t-1}, \dots, I_{t-d+1}) & if\ h = 1 \\ \hat{f}(\hat{I}_{t+h-1}, \dots, \hat{I}_{t+1}, I_t, \dots, I_{t-d+h}) & if\ h \in [2, \dots, d] \\ \hat{f}(\hat{I}_{t+h-1}, \dots, \hat{I}_{t+h-d}) & if\ h \in [d+1, \dots, H] \end{cases} \qquad (2)$$

Where $\hat{f}$ denotes the trained one-step-ahead model, and $h$ denotes the $h$th value in the output series, namely the $h$th step of the total horizon.

### 3.1.2 Direct strategy

Initially suggested by Cox[44], direct strategy constructs a set of models to forecast each step independently with the same observations as input variables. In other words, $H$ models are trained for each step respectively and it can be formulated as follows:

$$I_{t+h} = f_h(I_t, \dots, I_{t-d+1}) + w \qquad (3)$$

Where $h \in \{1, \dots, H\}$ and $w$ denotes the additive noise.

After the learning process, the estimation of the $H$ next values is given as follows:

$$\hat{I}_{t+h} = \hat{f}_h(I_t, I_{t-1}, \dots, I_{t-d+1}) \qquad (4)$$

Where $\hat{f}_h$ denotes the trained single-output model corresponding to the $h$th step.

### 3.1.3 MIMO strategy

In contrast to the above two strategies which model the data using single-output functions (see Eq. (1) and Eq. (3)), MIMO strategy is implemented using the function with multiple-input and multiple-output structure. The entire horizon is predicted using one multiple-output model in form of a vector of predicted values. MIMO strategy takes into account the existence of stochastic dependencies among future values, which is ignored in the iterated strategy and the direct strategy.

In the MIMO strategy, one multiple-output model $F$ is trained from the ILI rate time series $\{I_1, \dots, I_N\}$ as follows:
$$[I_{t+1}, \dots, I_{t+H}] = F(I_t, \dots, I_{t-d+1}) + \mathbf{w} \tag{5}$$

Where $F: \mathbb{R}^d \to \mathbb{R}^H$ denotes a vector-valued function, and $\mathbf{w} \in \mathbb{R}^H$ denotes the additive noise vector with a covariance that is not necessarily diagonal[45].

After the learning process, the estimation of the $H$ next values is given as follows:
$$[\hat{I}_{t+1}, \dots, \hat{I}_{t+H}] = \hat{F}(I_t, \dots, I_{t-d+1}) \tag{6}$$
Where $\hat{F}$ denotes the trained multiple-output model.

## 3.2 Models

### 3.2.1 Support vector regression (SVR)

The support vector regression (SVR) is a competitive approach deriving from the application of support vector machines for nonlinear regression and time series prediction problems. ε-SVR is adopted in this study for iterative strategy and direct strategy and briefly introduced as follows.

Given a training dataset $\{(x_i, y_i)\}_{i=1}^n \subset \mathbb{R}^d \times \mathbb{R}$, the goal of ε-SVR is to find a function $f(x)$ that has the deviation less than $\varepsilon$ from the observed $y_i$ for all the training data. The errors can be accepted if less than $\varepsilon$, while it is not allowed when they are larger than $\varepsilon$ [46]. $x_i$ denotes the d-dimensional input vector and $y_i$ denotes the corresponding target output. In the context of ILI rate forecasting, $x_i$ refers to d-dimensional previous observation of ILI rate series $\{I_t, \dots, I_{t-d+1}\}$; $y_i$ refers to the future value $I_{t+1}$. Given the form of $f(x)$ as Eq. (7), the formulation of ε-SVR optimization problem can be written in Eq. (8)

$$f(x) = \langle w, x \rangle + b, w \in \mathbb{R}^d, b \in \mathbb{R} \tag{7}$$

$$\begin{aligned}
&\text{minimize} \quad \frac{1}{2}\|w\|^2 + C\sum_{i=1}^{n}(\xi_i + \xi_i^*)\\
&\text{subject to} \quad \begin{cases} y_i - \langle w, x_i\rangle - b &\leq \varepsilon + \xi_i\\ \langle w, x_i\rangle + b - y_i &\leq \varepsilon + \xi_i^*\\ \xi_i, \xi_i^* &\geq 0 \end{cases}
\end{aligned} \qquad (8)$$

Where $\langle\cdot,\cdot\rangle$ denotes the dot product in the space $\mathbb{R}^d$; $\xi_i, \xi_i^*$ are slack variables relaxing the linear constraints for nonlinear regression data to cope with infeasible constraints of the optimization. Regularization parameter $C$ controls the trade-off between the variance and bias of the objective function. Moreover, kernel function $k(x, x')$ can be introduced to further solve nonlinear problems. As the most widely used kernel, Radial Basis Function kernel (RBF kernel, or Gaussian kernel) is adopted in this study, and the formulation of it is given as follows:

$$k(x, x') = exp(-\gamma\|x - x'\|^2) \qquad (9)$$

Where $\gamma = \frac{1}{2\sigma^2}$ is a parameter controlling the width of the kernel. In summary, three hyperparameters $\gamma, \varepsilon$ and $C$ need to be tuned during the training process.

Since SVR holds the structural risk minimization principle, namely minimizing an upper bound of the generalization error, it performs better in generalization than many other machine learning algorithms which adopt empirical risk minimization principle directly minimizing the training error[47,48]. Despite the mentioned merits of SVR, the standard formulation of SVR is inappropriate for multi-step-ahead forecasting with MIMO strategy on account of its inherent single-output structure[28]. The solution to this limitation of SVR was initially proposed by Pérez-Cruz et al.[49], designing a multi-dimensional SVR using a cost function with a hyper-spherical intensive zone. This new structure enables SVR to perform better in multi-dimensional forecasting tasks than predicting each dimension separately using standard SVR model. On this basis, Tuia et al.[50] constructed a multiple-output SVR model (MSVR) to estimate multiple biophysical parameters from remote sensing images simultaneously. Subsequent researches have explored and validated the effect of MSVR in diverse areas[50–52]. In this study, MSVR is applied as the implementation of MIMO strategy. Detailed information on MSVR can be found in [49,50].

### 3.2.2 Multilayer perceptron (MLP)

In general, Multi-Layer Perceptron (MLP), also called multi-layer feedforward neural network, comprises an input layer, one or more hidden layers consisting of activating neurons (nodes) and an output layer. In this study, a three-layer MLP with a single hidden layer is constructed. The input layer is formed by a set of neurons representing the input variables. Nodes in the hidden layer with appropriate nonlinear activation functions are used to cope with the information received by the input nodes. The output layer receives the calculation results from the hidden layer and provides output values as prediction results. Taking the MLP model for $H$-step-ahead ILI prediction with MIMO strategy as an instance, given a training dataset $\{(x_i, y_i)\}_{i=1}^{n} \subset \mathbb{R}^d \times \mathbb{R}^H$, the MLP model can be written as follows:

$$\boldsymbol{y} = \left( \sum_{j=1}^{k} w_{jh}^0 \sigma \left( \sum_{i=1}^{d} w_{ij}^r x_i + b_j^r \right) + b_h^0 \right), h = 1, \dots, H \tag{10}$$

Where $\boldsymbol{x}$ denotes the input vector, namely the previous lagged observations of ILI $\{I_t, \dots, I_{t-d+1}\}$; $\boldsymbol{y}$ denotes the output vector containing the future value of ILI series $\{I_{t+1}, \dots, I_{t+H}\}$; $\boldsymbol{w}^r, \boldsymbol{w}^0, \boldsymbol{b}^r$ and $\boldsymbol{b}^0$ respectively denote the weights and biases of hidden and output layers; $d, k$ and $H$ respectively denote the number of nodes in the input, hidden and output layer; $\sigma(\cdot)$ denotes the nonlinear activation function, which is Rectified Linear Unit function (ReLU) activation function in t this study.

Considering the structure of the output layer, MLP can be implemented with the MIMO strategy inherently. The hyperparameter to be optimized is the size of the hidden layer, namely the number of neurons in the hidden layer.

### 3.3 CLPSO-ML modeling framework

The model tuning process including feature selection and hyperparameters optimization always plays a significant role in the forecasting tasks with ML models. Nevertheless, these two subtasks are dominantly executed separately in literature, which is time consuming and ignores the interactive influences between features and hyperparameters [53]. Addressing these issues, the following adaptive scheme is developed for the multi-step-ahead influenza prediction.

### 3.3.1 CLPSO-based Framework for unified model tuning process

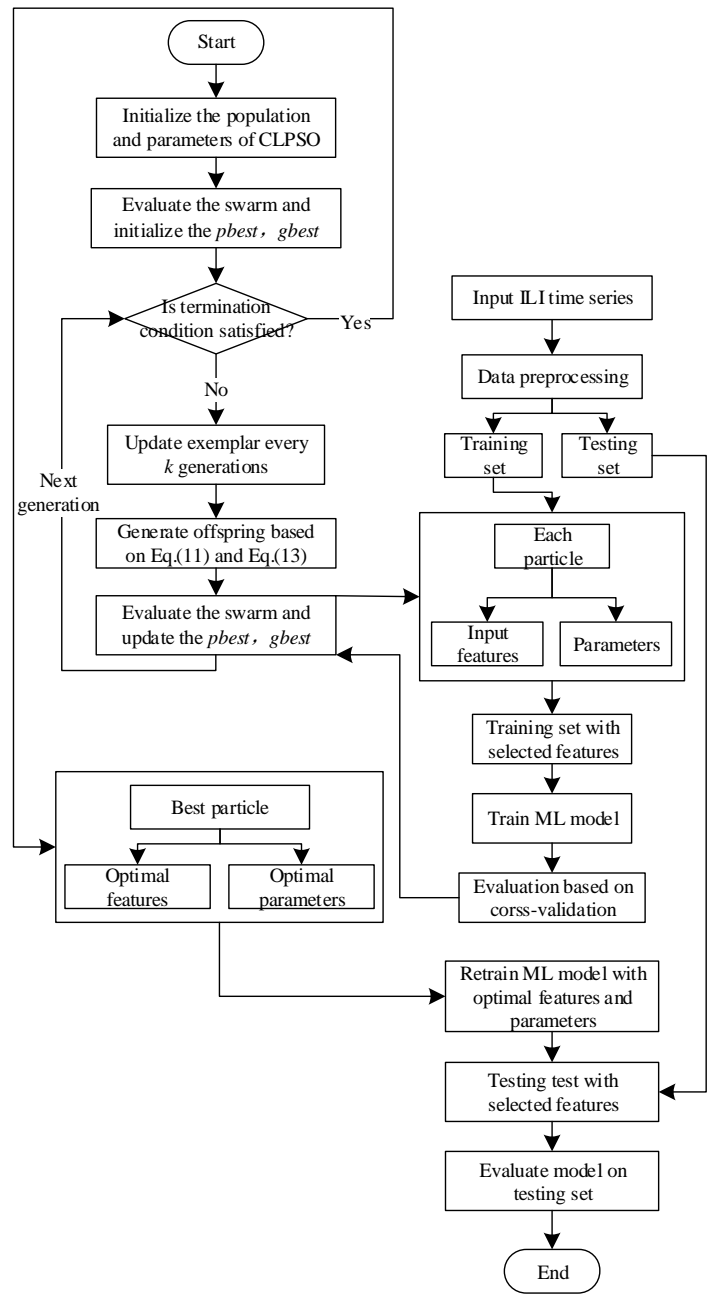

Fig 1 Flowchart of CLPSO based model tuning process for machine learning models

| | |
|---|---|
| 1 | **Start** |
| 2 | Calculate the particle dimension according to the number of candidate features and the length of encoded bit string of hyperparameters; |
| 3 | Initialize a population of particles with random positions; |
| 4 | Evaluate each particle with the fitness and initialize the *pbest* and *gbest*; |
| 5 | **While** the maximum iteration time or the maximum time of failing to improve the solution has not been reached: |
| 6 |     **For** each particle: |
| 7 |         **If** it's the first generation or the particle ceases improving for generations more than |
|    | the predefined refreshing gap:                                                                                           |
| 8  |     Assign the exemplar particle for each dimension according to the detailed rules in 3.3.3;        |
| 9  |     Update the velocity with Eq.(11) and the position with Eq.(13);                                  |
| 10 |     Transform the particle into the features and hyperparameters by decoding the binary string;      |
| 11 |     Train the ML model for multi-step-ahead influenza forecasting with the selected features and hyperparameters; |
| 12 |     Evaluate the particle taking mean squared error (MSE) of cross validation on the training set as fitness value; |
| 13 |     Update the personal best solution *pbest*;                                                       |
| 14 |   Update the global best solution *gbest*;                                                                     |
| 15 | Use the optimum features and hyperparameters decoded from *gbest* to retrain the prediction model on the whole training set; |
| 16 | Apply the trained model on the test set and assess the prediction performance under different strategies.                |
| 17 | **End**                                                                                                                  |

Fig 2 Pseudocode for CLPSO based model tuning process

The unified process of model tuning including hyperparameter optimization and feature selection for the adopted machine learning (ML) models using binary CLPSO algorithm is shown in Fig. 1. The left part of Fig. 1 demonstrates the executing procedure of the binary CLPSO optimization algorithm, which will be introduced in detail in Section 3.3.3. Each particle in the binary CLPSO algorithm represents candidate hyperparameters and input features of the ML models, which are encoded into a 0-1 series (See Section 3.3.2 for the specific encoding and decoding methods). The right part of Fig. 1 shows the training and testing process of the ML models. Firstly, ILI data are preprocessed and divided into training set and test set. During the training process, hyperparameters and selected features decoded from the particle are applied to train the model, and the performance of the trained model is evaluated as fitness value of the current particle. For each particle, the above training and evaluating process is repeated. In other words, the model is trained with different hyperparameters and features corresponding to each particle of every generation in CLPSO, until the termination condition is satisfied. After that, the best particle is reserved, comprising the optimal parameters and optimal selected features which will be subsequently exploited to retrain the influenza prediction model on the whole training set. Finally, the model with multi-step strategies are applied on the test set, and metrics of forecasting performance are calculated for the comparison of different strategies with specific ML model. Pseudocode of the above procedure is illustrated in Fig.2.

### 3.3.2 Binary particle representation of features and parameters

In order to make CLPSO available for the unified optimization of feature selection and hyperparameter tuning, a binary encoding approach is designed to contain information of selected features and parameters in each particle. The particle in the binary CLPSO consists of two parts: the feature masks, and hyperparameters of the adopted ML models, specifically SVR and MLP in this context. In terms of the feature mask part, the 0-1 string with length of $d$ correspond to the d-dimensional input features $\{I_t, \ldots, I_{t-d+1}\}$, namely the lagged values of ILI. The value 1 denotes that the feature at the corresponding position is selected, while value 0 means the corresponding feature is excluded. Considering the parameter part, a list of candidate values is predefined for each hyperparameter involved in the ML model. The candidate values are sorted and each corresponds to a unique integer index in the list. In contrast to the feature masks, the 0-1 string in this part works as a whole, representing the index of the value in the candidate list by converting the binary string to decimal integer. Fig. 3 shows an intuitive example of the binary particle representation for SVR.

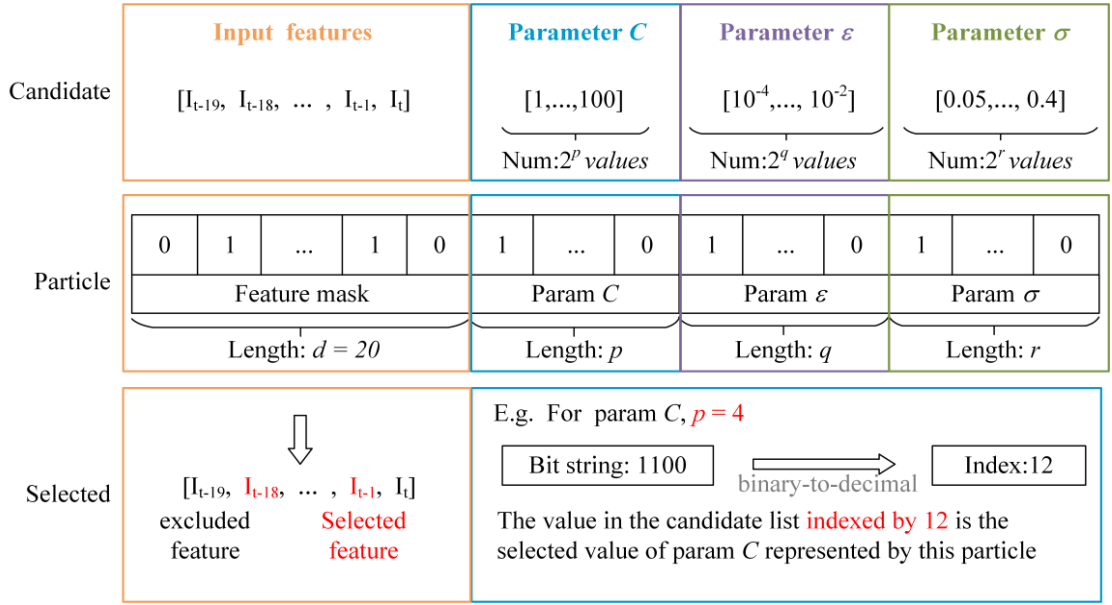

Fig 3 Binary representation of a particle

### 3.3.3 Binary CLPSO Algorithm

This section briefly introduces the used binary CLPSO, and more detailed information about the original CLPSO algorithm can be found in [41]. CLPSO is a variant of PSO using a comprehensive learning strategy that the best position of other particle is applied to update the velocity of each particle. Population diversity is preserved and the ability of global search is improved under such mechanism. The velocity is update with the following equation:

$$v_i^m = w \times v_i^m + c \times r_i^m \times \left(pbest_{fi(d)}^m - x_i^m\right) \quad (11)$$

where $v_i^m$ denotes the velocity and $x_i^m$ denotes the position of the $i$th particle in the $m$th dimension; $c \in \mathbb{R}$ is the acceleration coefficient; $w$ is called inertia weight which is used to balance the global and local search abilities; $r_i^m$ is a random number in the range [0,1]; and $pbest_{fi(m)}^m$ refers to the $m$th dimension of the exemplar particle's best previous position, with the index of exemplar $fi(m)$ selected according to a predefined learning probability $Pc_i$. the learning probability is empirically calculated as follows:

$$Pc_i = 0.05 + 0.45 \times \frac{\exp\left(\frac{10(i-1)}{ps-1}\right) - 1}{\exp(10) - 1} \quad (12)$$

where $ps$ denotes size of swarm. Specifically, the index $fi(m)$ of the exemplar particle is determined as follows. For each dimension of particle $i$, if a generated random number in the range [0,1] is greater than probability $Pc_i$, the corresponding dimension will learn from its own, namely $fi(m) = i$; otherwise, it will learn from another particle. To select another exemplar particle, the fitness values of two randomly selected particles are compared to find the smaller one as the exemplar, excluding the particle whose velocity is updated. If all the exemplars of each dimension are the particle itself, then a randomly selected dimension is forced to learn from the corresponding dimension of another particle's best previous position. Once the all exemplars of a particle have been selected, the particle is allowed to learn from these exemplars until the particle is not improved for $k$ generations ($k$ is called the refreshing gap).

In this study, the position of particle is expressed as a binary bit vector, and thereby the velocity is the probability for each dimension to change from 1 to 0 or inversely. The update equation of the position is as follows:

$$x_i^m = \begin{cases} 1 & \text{if rand} < S(v_i^m) \\ 0 & \text{otherwise} \end{cases},$$

$$S(v_i^m) = \frac{1}{1 + \exp(-v_i^m)} \quad (13)$$

Where rand is a uniformly distributed random number in [0,1], and S(·) is a sigmoid transfer function.

Theoretically, the computational complexity of the binary CLPSO in the worst case is: $O(ps \times T \times M)$, where $ps$ denotes size of swarm, $T$ denotes the maximum iteration times, and $M$ denotes the length of particle dimension. Among the three determinants, $ps$ and $T$ are merely related to the parameter setting of CLPSO algorithm, while $M$ is determined by how many week lags of ILI rate is taken as candidate features and how many candidate values are predefined for hyperparameters. In order to ensure the fair comparison between different multi-step-ahead strategies, the number of week lags, namely the dimension of candidate features is fixed in the following experiment. On the other hand, the representation of hyperparameters in particles induces a logarithmic relationship (with the base = 2) between the length of bit string and the total number of candidate values for all the hyperparameters, which makes $M$ increase slowly when there are many hyperparameters in a ML model or the candidate value list is long. Therefore, the computational complexity of the total framework will be acceptable in this study.

# 4. Experiment design

## 4.1 Data source and preprocessing

Weekly influenza like illness (ILI) rates in the Southern and Northern China from the first week of 2010 to the 12th week of 2020 are collected from Influenza Weekly Report issued on the website of Chinese National Influenza Center. The issued ILI rate is calculated by the following formula:

$$\text{ILI rate} = \frac{the\ number\ of\ weekly\ ILI\ hospital\ visits}{the\ number\ of\ total\ weekly\ hospital\ visits} \quad (14)$$

the Northern China ILI rate series (abbreviated as N_ILI) and the Southern China ILI rate series (abbreviated as S_ILI) are showed as follows.

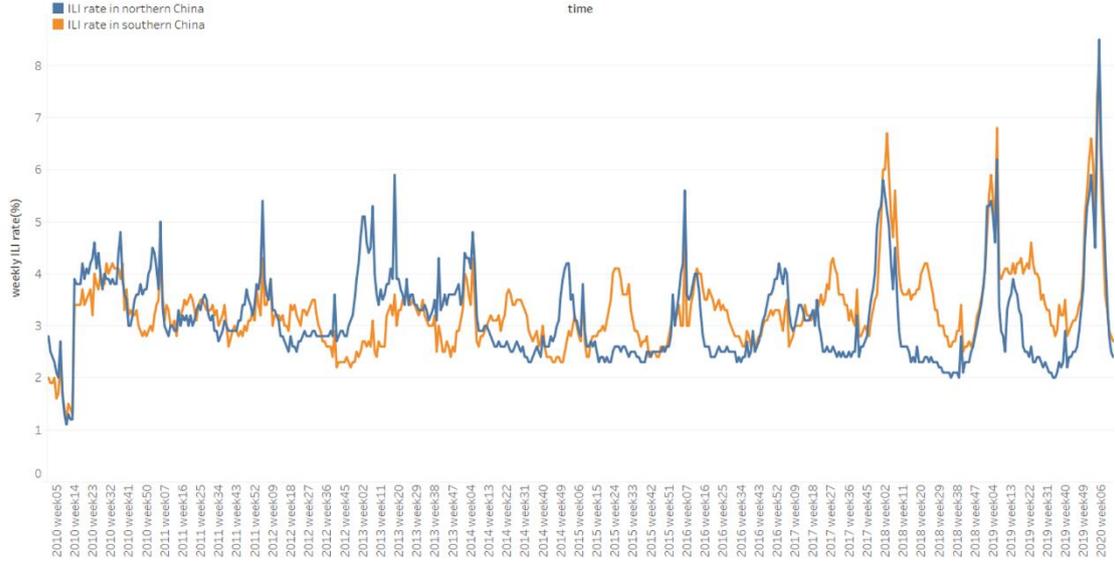
Fig 4 ILI rates time series in southern and northern China

Both N_ILI and S_ILI have a missing value at the 40th week of 2011. Considering the trend of the series, the mean of the values at the subsequent and the precedent point is calculated to fill the missing value. Logarithmic transformation is applied to both ILI rate series with log base 10, which could make the size of the seasonal variation same across the whole series and therefore make the forecasting model simpler [54]. The transformation is reversed to achieve the predicted values of the original scale.

## 4.2 Experimental procedure

Three multi-step-ahead strategies are applied in the experiment, namely iterated strategy, direct strategy, and MIMO strategy. Considering the inherent single-output property, the standard SVR can only be implemented with the iterated strategy and the direct strategy. When it comes to MIMO strategy, MSVR is adopted. Whereas MLP fits all the strategies since it allows multiple output inherently.

Fig 5 shows the whole experimental procedure. Data preprocessing including missing value imputation and logarithmic transformation is conducted firstly. Then for each prediction horizon (in this case the horizon ranges from 2 to 10), N_ILI and S_ILI series are respectively converted to the form of input variables $x$ and prediction target $y$, where $x = [I_t, I_{t-1}, \ldots, I_{t-d+1}] \in \mathbb{R}^d$ represents the input samples of observed ILI rates from the week $t-d+1$ to the week $t$, and $y = [I_{t+1}, I_{t+2}, \ldots, I_{t+H}] \in \mathbb{R}^H$ represents the true values of the ILI rates as prediction target from the week $t+1$ to the week $t+H$, with $H$ as the forecast horizon. Then $x$ and $y$ are split into the training samples including the first 2/3 of the dataset, and the test samples containing the last 1/3 of the dataset. After the datasets are prepared, the model tuning process including feature selection and hyperparameter optimization is performed, and the final trained model is employed on the test dataset with the three different multi-step strategies. Afterward, four accuracy metrics including both statistical metrics and influenza outbreak metrics are computed. The above procedure is repeated 20 times for each horizon, and finally the performance of each model for each prediction horizon is assessed by the mean of the aforementioned metrics. Friedman test was conducted to test if the means of performance measures are statistically different among the different models with different strategy for each prediction horizon. If so, Nemenyi test was subsequently employed as post-hoc test to identify the significantly different prediction models in multiple pair wise comparisons at the significance level of 0.05. Evaluation metrics are specifically illustrated in Section 4.3, and implementation details of CLPSO-ML modeling framework are presented in Section 4.4

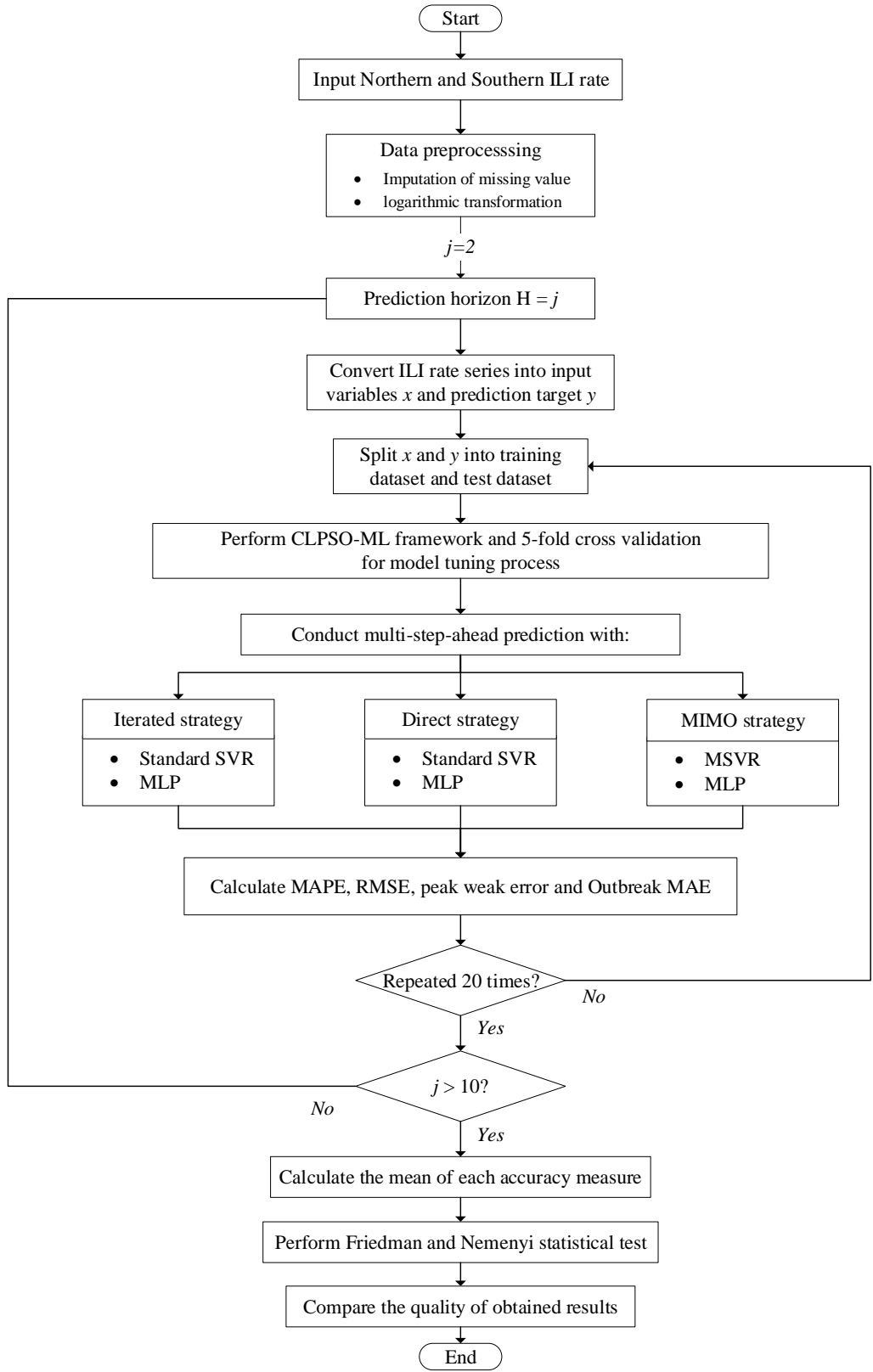

Fig 5 Experiment procedure for ILI multi-step-ahead forecasting

## 4.3 Evaluation of model performance

### 4.3.1 Statistical metrics

To assess the forecasting performance of different model, two statistical metrics, MAPE (Mean Absolute Percentage Error) and RMSE (Root Mean Square Error), are used in the experiment. MAPE is scale independent and hence is frequently adopted to evaluate forecasting performance across different datasets. Nevertheless, its disadvantage lies on the heavier penalty on positive errors than on negative errors. Therefore, RMSE is also introduced to measure the accuracy of prediction. As a scale-dependent metric, RMSE can only be compared among different models for the same dataset [28]. The definitions of MAPE and RMSE are given as follows:

$$MAPE = \frac{1}{n}\sum_{j=1}^{n}\left|\frac{I_j - \hat{I}_j}{I_j}\right| \qquad (15)$$

$$RMSE = \sqrt{\frac{1}{n}\sum_{j=1}^{n}(I_j - \hat{I}_j)^2} \qquad (16)$$

Where $m$ denotes the number of fitted ILI rates, $I_j$ denotes the $j$th observed ILI rate, and $\hat{I}_j$ denotes the $j$th predicted ILI rate.

### 4.3.2 Influenza outbreak metrics

Besides the statistical metrics, in the context of influenza prediction, it is of great importance to accurately predict the situation during influenza outbreak periods. Before giving the definition of the metrics, it is necessary to define the influenza outbreak seasons of the whole ILI time series. As is discussed earlier, the specific rules mentioned in [14,15], which can be used to judge whether the ILI value of each time point belongs to influenza outbreak period or not, are not suitable for the adopted China weekly ILI rates. Therefore the outbreak periods in this study are roughly defined as the winter months from the 45[th] week in this year to the 8[th] week in the next year, referring to the description of Chinese National Influenza Center. In this case, two influenza outbreak metrics, the Peak Week Error (PWE) and the Outbreak MAE, are employed to assess the forecasting performance during influenza outbreak periods.

For each specific outbreak season, an influenza peak is defined as the week of highest ILI rate in the outbreak period. the peak week error (PWE) of a given outbreak season is the absolute difference between the observed peak week and the forecasted peak week [17], as given by

$$PWE = |t_p - \hat{t}_p| \qquad (17)$$

Where $t_p$ denotes the observed week of the peak and $\hat{t}_p$ denotes the predicted week of the peak.

PWE is aimed at the single peak point of a given influenza outbreak season, while the Outbreak MAE is used to evaluate the forecasting performance across the whole outbreak period. Outbreak MAE is the abbreviation of the mean absolute error of a consecutive influenza outbreak period, as given by

$$Outbreak\ MAE = \frac{1}{s}\sum_{j=1}^{s}|I_j^o - \hat{I}_j^o| \qquad (18)$$

Where $s$ denotes the number of ILI rate values encompassed in a given outbreak season; $I_j^o$ denotes the $j$th observed ILI rate in the given outbreak season, and $\hat{I}_j^o$ denotes the $j$th predicted ILI rate in the given outbreak season.

Three influenza outbreak seasons are included in the test samples of N_ILI and S_ILI. Fig. 6 shows the outbreak periods and the peak point of each period on N_ILI and S_ILI. For N_ILI, three peak values appear at the 1st week of 2018, the 6th week of 2019 and the 5th week of 2020, while the peak weeks of S_ILI are the 3rd week of 2018, the 6th week of 2019 and 5th week of 2020. The PWE and Outbreak MAE on the test sets are the averaged value of the three outbreak periods.

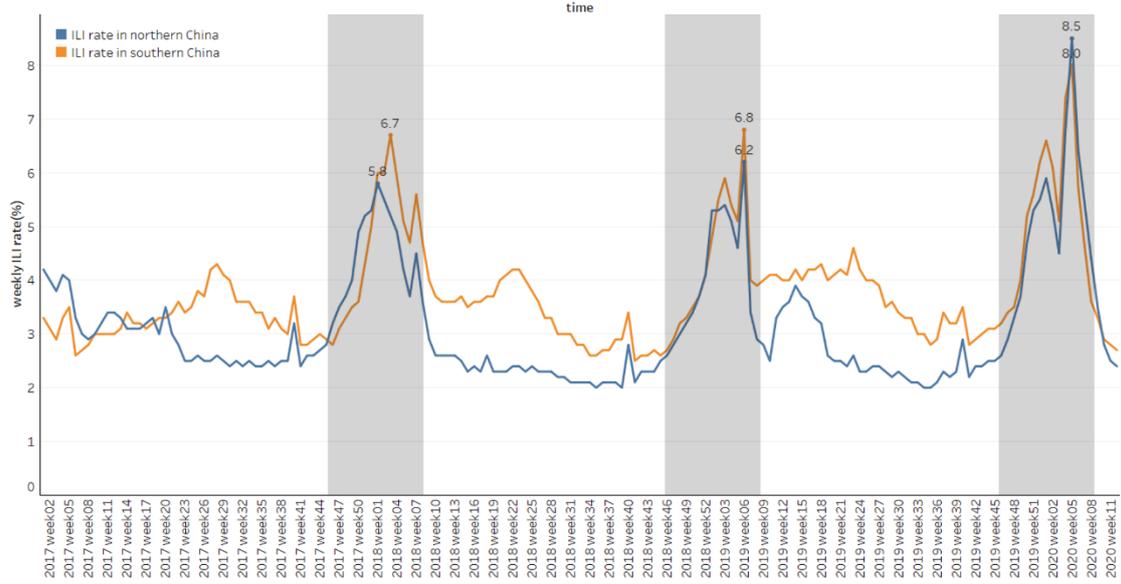

Fig 6 Influenza outbreak seasons and peaks in the test sets of N_ILI and S_ILI

## 4.4  Implementation of ML models and CLPSO algorithm

*Scikit-learn v0.24.1* package of python is employed to build the standard RBF kernel SVR models taking iterated and direct strategies and MLP models taking all the three strategies. MSVR is rewritten in python according to the MATLAB version of program provided by Tuia et al [50]. Table 2 demonstrates the space of hyperparameters for SVR and MLP.

Table 2 parameter space for optimization in SVR and MLP

| ML model | Parameter | Parameter Space |
|---|---|---|
| SVR | C | a geometric sequence with 16 numbers starting from 1 and ending with 100 |
| | ε (epsilon) | a geometric sequence with 4 numbers starting from $10^{-4}$ and ending with $10^{-2}$ |
| | γ (gamma) | [0.05, 0.1, 0.2, 0.4] |
| MLP | size of hidden layer | [10,20,50,100] |

ILI rates of previous 20 weeks are selected as the candidate input variables, i.e., $d = 20$. There are also some adjustable parameters in the CLPSO algorithm, which are determined in a fashion of trail-and-error considering the trade-off between prediction accuracy and computational time. Specifically, the termination condition is that the number of total iterations exceeds 200 or the global fitness value does not improve for 30 consecutive iterations; the acceleration coefficient ($c$) is set to 2; the inertia weight decreases linearly from 0.9 to 0.4 with the increase of iterations; the number of particles in the swarm is set to 8; the refreshing gap $k$ is set to 8.

## 5. Results and discussion

The prediction performances of the two adopted machine learning models (SVR and MLP) across the three multi-step strategies (Iter, Dir and MIMO, as the abbreviation of the Iterated strategy, the Direct strategy and the MIMO strategy respectively in the following tables) in terms of two statistical metrics MAPE, RMSE and two influenza outbreak metrics PWE, Outbreak MAE over the prediction horizon ranging from 2 to 10 on the two weekly ILI rate series (i.e. N_ILI and S_ILI) are demonstrated in Table 3 and Table 4 respectively. Fig. 7 and 9 show the performance of the different multi-step strategies for the same machine learning model more intuitively, with the increase of prediction horizon on the two ILI rate series. Each row of the figures corresponds to a ML model, with the first row for SVR and the second row for MLP. Four columns of the figures depict the four adopted accuracy metrics respectively, namely MAPE, RMSE, PWE and Outbreak MAE. In Fig. 8 and 10, we choose a typical horizon (horizon = 4) to depict the prediction results of the three strategies on test sets of N_ILI and S_ILI. Each row of the figures corresponds to a ML model, with the first row for SVR and the second row for MLP; Four columns of the figures respectively depict the 1-, 2-, 3-, and 4-step-ahead prediction results.

Table 3 Prediction accuracy metrics for weekly ILI rate in Northern China.

| Metric | Model | Forecast horizon (H) | | | | | | | | |
|---|---|---|---|---|---|---|---|---|---|---|
| | | 2 | 3 | 4 | 5 | 6 | 7 | 8 | 9 | 10 |
| MAPE | SVR-Iter | **0.0872** | **0.0977** | 0.1104 | 0.1173 | **0.1230** | 0.1318 | 0.1410 | 0.1473 | 0.1559 |
| | SVR-Dir | 0.0890 | 0.1028 | 0.1149 | 0.1233 | 0.1324 | 0.1403 | 0.1479 | 0.1552 | 0.1609 |
| | SVR-MIMO | 0.0905 | 0.1012 | **0.1099** | **0.1168** | 0.1242 | **0.1309** | **0.1374** | **0.1425** | **0.1465** |
| | MLP-Iter | 0.0995 | 0.1122 | 0.1231 | 0.1369 | 0.1449 | 0.1496 | 0.1533 | 0.1593 | 0.1641 |
| | MLP-Dir | 0.0982 | 0.1091 | 0.1186 | 0.1291 | 0.1390 | 0.1463 | 0.1520 | 0.1594 | 0.1693 |
| | MLP-MIMO | 0.0974 | 0.1060 | 0.1203 | 0.1235 | 0.1392 | 0.1414 | 0.1476 | 0.1559 | 0.1559 |
| RMSE | SVR-Iter | 0.6008 | **0.6341** | 0.6961 | 0.7006 | **0.7305** | **0.7660** | 0.7983 | 0.8068 | 0.8464 |
| | SVR-Dir | 0.5936 | 0.6403 | 0.7083 | 0.7430 | 0.7831 | 0.8203 | 0.8589 | 0.8868 | 0.9071 |
| | SVR-MIMO | 0.5902 | 0.6355 | **0.6762** | **0.6987** | 0.7306 | 0.7689 | **0.7914** | **0.8065** | **0.8180** |
| | MLP-Iter | **0.5789** | 0.6385 | 0.6869 | 0.7498 | 0.7784 | 0.8050 | 0.8197 | 0.8450 | 0.8615 |
| | MLP-Dir | 0.6023 | 0.6554 | 0.6993 | 0.7449 | 0.7799 | 0.8106 | 0.8387 | 0.8671 | 0.9044 |
| | MLP-MIMO | 0.6163 | 0.6561 | 0.6911 | 0.7156 | 0.7601 | 0.7845 | 0.8061 | 0.8433 | 0.8415 |
| PWE | SVR-Iter | 2.1000 | 2.0667 | **<u>1.9417</u>** | **<u>1.7267</u>** | **<u>1.7389</u>** | **<u>1.6762</u>** | **<u>1.7979</u>** | **<u>1.8352</u>** | **<u>1.9700</u>** |
| | SVR-Dir | 2.0667 | **<u>1.8611</u>** | 2.1333 | 2.1533 | 2.1083 | 2.4786 | 2.9021 | 3.2519 | 3.7500 |
| | SVR-MIMO | 2.3167 | 2.1056 | 1.9458 | 1.7967 | 1.7806 | 1.7690 | 1.8354 | 2.0056 | 2.2217 |
| | MLP-Iter | **<u>1.5417</u>** | 1.9111 | 2.3000 | 2.8167 | 3.2111 | 3.5095 | 3.0688 | 3.4907 | 3.5700 |
| | MLP-Dir | 1.8250 | 1.9722 | 2.5542 | 2.8867 | 3.9194 | 3.8881 | 4.1083 | 4.6463 | 5.2017 |
| | MLP-MIMO | 2.2250 | 2.3611 | 2.3917 | 2.7000 | 3.1778 | 4.2000 | 4.1958 | 4.3481 | 4.2667 |
| Outbreak MAE | SVR-Iter | **<u>0.6900</u>** | **<u>0.7404</u>** | 0.8522 | **<u>0.8693</u>** | 0.9382 | 1.0107 | 1.0984 | 1.1507 | 1.2532 |
| | SVR-Dir | 0.7069 | 0.7773 | 0.8811 | 0.9463 | 1.0329 | 1.1079 | 1.1982 | 1.2772 | 1.3494 |
| | SVR-MIMO | 0.7116 | 0.7743 | 0.8412 | 0.8839 | 0.9556 | 1.0452 | 1.1066 | 1.1687 | 1.2312 |
| | MLP-Iter | 0.7072 | 0.7730 | 0.8357 | 0.9250 | 0.9825 | 1.0570 | 1.1151 | 1.1883 | 1.2551 |
| | MLP-Dir | 0.7383 | 0.7938 | 0.8500 | 0.9160 | 0.9620 | 1.0316 | 1.1028 | 1.1491 | 1.2204 |
| | MLP-MIMO | 0.7538 | 0.7988 | **<u>0.8282</u>** | 0.8717 | **<u>0.9335</u>** | **<u>1.0025</u>** | **<u>1.0597</u>** | **<u>1.1476</u>** | **<u>1.2122</u>** |

Note: the smallest value of each metric is underlined and set in boldface

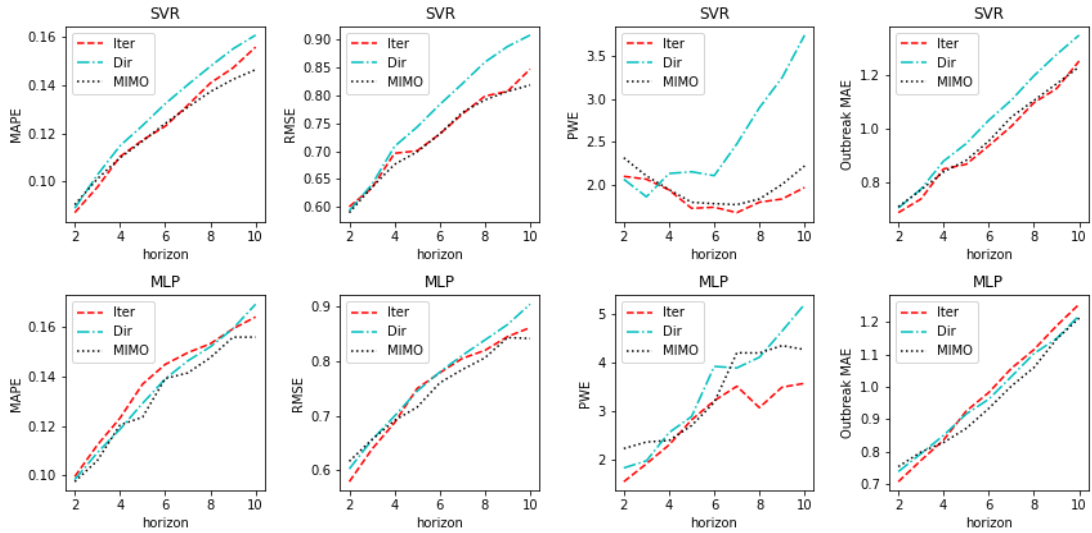

Fig 7 Comparisons between the three multi-step strategies over different horizons for N_ILI

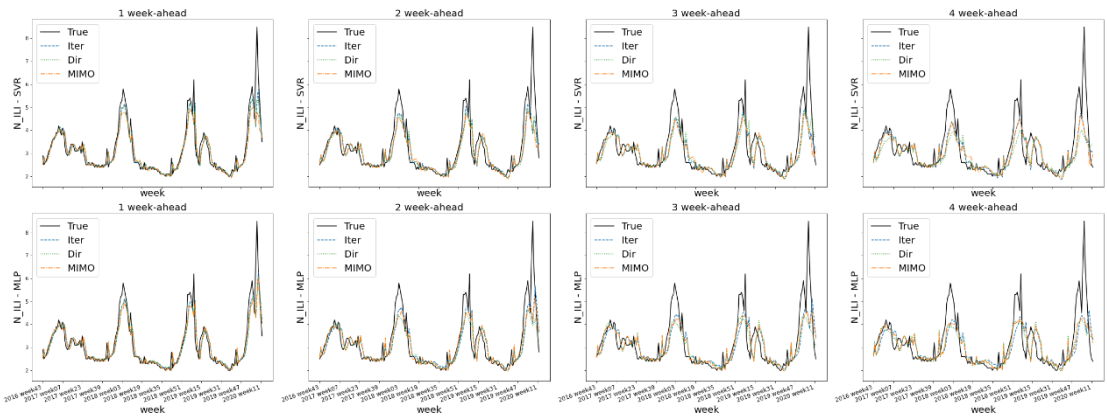

Fig. 8 Prediction results of the three strategies for N_ILI (horizon = 4)

Considering the presented results on N_ILI, several observations can be drawn from Table 3 and Fig. 7:

- As per MAPE, MIMO strategy performs best implemented with both SVR and MLP generally, and most of the SVR models perform better than MLP models. Specifically, for SVR, when the horizon(H) equals to 2,3 and 6, the top one is iterated strategy, while MIMO strategy keeps the best over all of other horizons. While considering the statistical test, there are no significant difference (at the 0.05 level, the same hereinafter) between MIMO and iterated strategy when H = 3-9. The direct strategy significantly performs worse than the other two strategies over most of the horizons. For MLP, MIMO strategy derives the best performance over all the horizons except for H = 4 and 6, where the direct strategy ranks first. While the significant differences between MIMO and direct strategy with MLP only appears when H = 10. Iterated strategy ranks last of the three in most of the cases.
- As per RMSE, MIMO strategy performs best with SVR over most of the horizons and with MLP over the horizons larger than 4. The iterated strategy is also competitive while being implemented with SVR, and ranks the first with MLP when H $\leqslant$ 4. In particular, for SVR, iterated strategy is the top when H = 3, 6 and 7, and MIMO strategy keeps the first for the rest. However, no significant distinction between the above two strategies over all the horizons involved according to the Nemenyi test. Direct strategy performs significantly worst when H = 4-10. For MLP, iterated strategy ranks first when H = 2-4, and MIMO strategy derives the smallest RMSE on the rest horizons. Statistical test shows that the RMSE of the MIMO strategy is significantly smaller than the iterated strategy at H = 5, and significantly larger at H = 2. The direct strategy performs significantly worse than the other two when H = 8 and 10.
- As per PWE, iterated strategy derives the smallest week difference between the predicted peak week and the true peak week on average over most of the horizons with both SVR and MLP. Meanwhile, SVR models perform better than MLP models over the majority of horizons in general. Specifically, for SVR, the iterated strategy outperforms the other two over each horizon except for H = 2 and 3, where the direct strategy acts best. According to Nemenyi test, the PWE of the direct strategy is significantly larger when H = 7-10, and MIMO strategy performs significantly poorer than the direct strategy at H = 3. For MLP, iterated strategy ranks first apart from H = 5 and 6, where MIMO is the top. The iterated strategy significantly outperforms MIMO strategy at H = 2 and significantly beats the direct strategy when H = 8 and 10.

- As per Outbreak MAE, the iterated strategy performs best during influenza outbreak periods over most of the horizons with SVR and at H = 2, 3 with MLP, while MIMO strategy performs best in other cases. In particular, for SVR, iterated strategy derives the smallest Outbreak MAE over each horizon except for H = 4 and 10, where MIMO strategy is the top. While there is no significant difference between iterated strategy and MIMO strategy for all horizons. The direct strategy ranks last and significantly worse than the other two when H > 2. For MLP, MIMO strategy has the best performance on Outbreak MAE when H > 3. The iterated strategy ranks first over H = 2 and 3, but becomes the last when H > 4. Significant distinction appears between iterated strategy and MIMO strategy when H = 5 and 8.

Overall, findings on ILI rates of the Northern China can be summarized as follows:
1) MIMO strategy derives the least errors for the majority of prediction horizons with both ML models according to the two statistical metrics MAPE, RMSE and one of the influenza outbreak metrics Outbreak MAE.
2) Iterated strategy is also competitive considering the above metrics since there is no significance between iterated and MIMO strategy in many cases, and even better especially for short length of horizons (H = 2,3). More importantly, the iterated strategy ranks the first over most of the horizons by PWE.
3) Comprehensively considering all the metrics, the horizon of 4 seems a turning point, where iterated strategy has an edge for shorter length of horizon (H = 2 and 3), while MIMO strategy is potentially more adaptive for longer horizon (H ≥ 4).

Table 4 Prediction accuracy metrics for weekly ILI rate in Southern China

| Metric | Model | Forecast horizon (H) | | | | | | | | |
|---|---|---|---|---|---|---|---|---|---|---|
| | | 2 | 3 | 4 | 5 | 6 | 7 | 8 | 9 | 10 |
| MAPE | SVR-Iter | 0.0912 | 0.1015 | 0.1111 | 0.1180 | 0.1238 | 0.1279 | 0.1325 | 0.1350 | 0.1385 |
| | SVR-Dir | 0.0904 | 0.1011 | 0.1098 | 0.1169 | 0.1230 | 0.1274 | 0.1312 | 0.1342 | 0.1379 |
| | SVR-MIMO | 0.0929 | 0.1006 | **0.1090** | **0.1152** | 0.1216 | **0.1253** | 0.1320 | **0.1333** | 0.1394 |
| | MLP-Iter | 0.0909 | 0.1031 | 0.1103 | 0.1173 | 0.1226 | 0.1267 | 0.1300 | 0.1335 | 0.1367 |
| | MLP-Dir | 0.0919 | 0.1019 | 0.1111 | 0.1185 | 0.1253 | 0.1294 | 0.1328 | 0.1374 | 0.1404 |
| | MLP-MIMO | **0.0902** | **0.0994** | 0.1096 | 0.1166 | **0.1215** | 0.1283 | **0.1295** | 0.1358 | **0.1353** |
| RMSE | SVR-Iter | 0.6102 | 0.6738 | 0.7452 | 0.7991 | 0.8455 | 0.8876 | 0.9259 | 0.9496 | 0.9774 |
| | SVR-Dir | 0.6055 | 0.6802 | 0.7494 | 0.7996 | 0.8465 | 0.8851 | 0.9239 | 0.9487 | 0.9727 |
| | SVR-MIMO | 0.6715 | 0.7080 | 0.7439 | 0.7944 | 0.8455 | 0.8775 | 0.9325 | 0.9436 | 0.9798 |

|  | MLP-Iter | 0.6000 | 0.6676 | **0.7124** | **0.7777** | 0.8262 | **0.8713** | **0.9036** | **0.9339** | 0.9611 |
|  | MLP-Dir | 0.5999 | 0.6670 | 0.7262 | 0.7887 | 0.8433 | 0.8803 | 0.9176 | 0.9401 | 0.9637 |
|  | MLP-MIMO | **0.5943** | **0.6598** | 0.7188 | 0.7794 | **0.8255** | 0.8776 | 0.9058 | 0.9550 | **0.9508** |
| PWE | SVR-Iter | 1.4417 | 1.8778 | 2.6875 | 2.6600 | 3.5861 | 3.5738 | 3.3958 | 3.5296 | 4.1467 |
|  | SVR-Dir | 1.4583 | 2.5722 | 3.5750 | 3.5967 | 4.3917 | 4.3262 | 5.0813 | 4.7463 | 5.9517 |
|  | SVR-MIMO | 1.5667 | **1.5833** | **1.8667** | **1.9467** | **1.9833** | **2.4095** | **2.4333** | **2.9185** | **3.1517** |
|  | MLP-Iter | **1.4167** | 1.8333 | 2.1417 | 2.4700 | 2.7417 | 3.3000 | 3.0708 | 3.6778 | 4.1017 |
|  | MLP-Dir | 1.4583 | 1.7722 | 2.8500 | 3.8133 | 4.9750 | 5.6024 | 6.1292 | 7.1630 | 8.0933 |
|  | MLP-MIMO | 1.4500 | 1.7500 | 2.3458 | 2.8500 | 3.3694 | 3.9190 | 4.0688 | 4.8722 | 4.7383 |
| Outbreak MAE | SVR-Iter | 0.7945 | 0.8777 | 0.9699 | 1.0395 | 1.1122 | 1.1912 | 1.2699 | 1.3417 | 1.4285 |
|  | SVR-Dir | 0.7931 | 0.8900 | 0.9797 | 1.0473 | 1.1177 | 1.1877 | 1.2707 | 1.3371 | 1.4147 |
|  | SVR-MIMO | 0.8748 | 0.9219 | 0.9668 | 1.0365 | 1.1123 | 1.1764 | 1.2767 | 1.3261 | 1.4097 |
|  | MLP-Iter | 0.7890 | 0.8647 | **0.9195** | **1.0011** | **1.0750** | **1.1593** | **1.2296** | 1.3108 | 1.3907 |
|  | MLP-Dir | 0.7943 | 0.8713 | 0.9346 | 1.0134 | 1.0914 | 1.1626 | 1.2333 | **1.3011** | **1.3789** |
|  | MLP-MIMO | **0.7848** | **0.8629** | 0.9267 | 1.0074 | 1.0834 | 1.1602 | 1.2433 | 1.3464 | 1.3803 |

Note: the smallest value of each metric is underlined and set in boldface

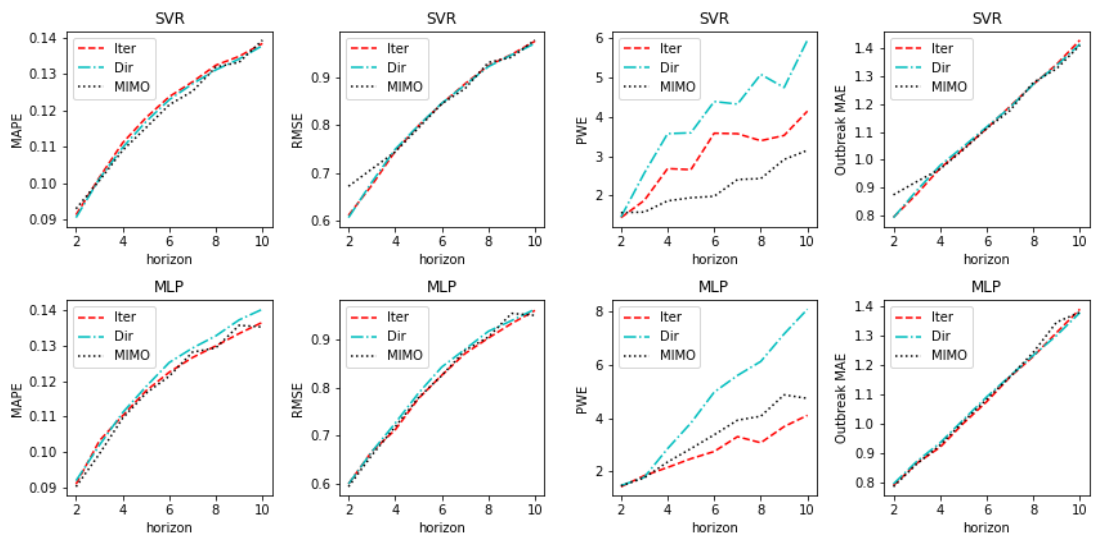

Fig 9 Comparisons between three multi-step strategies over different horizons for S_ILI

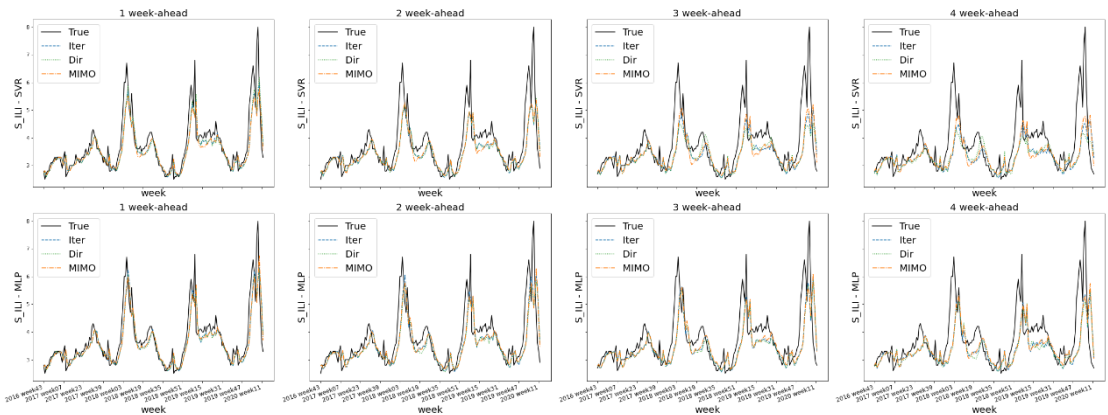

Fig. 10 Prediction results of the three strategies for S_ILI (horizon = 4)

As per the results in the case of S_ILI, one can deduce the following observation from Table 4 and Fig.9:

- As per MAPE, MIMO strategy performs slightly better than the other two over most of the horizons. Specifically, for SVR, direct strategy ranks the first when H = 2, 8 and 10, with MIMO strategy as the best over the rest horizons. According to the statistical test, MIMO strategy significantly outperforms the others when H = 5-7 and 9, while there is no significant difference between MIMO strategy and direct strategy when H = 3,4 and 8. The MAPE of iterated strategy is significantly larger compared with the other two strategies when H = 4-9, and larger than direct strategy at H = 2. For MLP, MIMO strategy derives the best performance over all the horizons except for H = 7 and 9, where iterated strategy is the top. While MIMO strategy and iterated are significantly distinct only when H = 3,7 and 9 according to the Nemenyi test. The direct strategy ranks last of the three significantly in most of the cases.
- As per RMSE, generally these three strategies perform similarly and none of strategies obviously win on the majority of horizons. While most MLP models performs better than SVR models only except for H = 9. In particular, for SVR, MIMO strategy ranks the first when H = 4,5,7 and 9, direct strategy becomes the top when H = 2,8,10, and iterated strategy is the best when H = 3 and 6. The only significant difference appears between MIMO strategy and the rest when H < 4. For MLP, MIMO slightly outperforms the others when H = 2,3,6 and 10, while iterated strategy outperforms a little over the rest horizon. The RMSE of direct strategy keeps the largest when H = 4-8 and 10, although the three of the strategies have no significant differences over the majority of horizons.
- As per PWE, MIMO strategy implemented with SVR ranks the first obviously on most of the horizons, while iterated strategy combining with MLP derives the best performance for all horizons except H = 3. Specifically, for SVR, MIMO strategy performs best over all the horizons except H = 2, where iterated strategy becomes the first. And statistical tests show that MIMO significantly outperforms the other two on the overwhelming majority of horizons. For MLP, iterated strategy is the top over all the horizons apart from H = 3, where MIMO strategy wins. Nevertheless, no significant distinction exists between MIMO strategy and iterated strategy according to the Nemenyi test, though the two averaged PWE value of 20 repeated experiments seems to have relatively large gap. The direct strategy significantly underperforms the others when H > 3.

- As per Outbreak MAE, the situation is similar with that of RMSE, where none of the strategies obviously outperforms the others over most of the horizons, and MLP models beat the SVR models in most of the cases. In particular, for SVR, MIMO strategy ranks the first when H = 4,5,7,9 and 10, iterated strategy becomes the top when H = 3,6,7, and direct strategy is the best when H = 2. The only significant difference appears between direct strategy and iterated strategy when H = 2. For MLP, iterated strategy performs best when H = 4-8, MIMO strategy becomes the first when H = 2,3, and direct strategy is the top when H = 9,10. There are no significant distinctions among the three strategies in most of the cases.

In summary, the distinction between the performance of three modelling strategies are smaller on the Southern China ILI rates than the Northern China except for PWE. Despite of this, there is some common ground of the results for the both ILI series.

1) MIMO strategy can be identified as the most competitive multi-step-ahead modelling strategy, comprehensively considering all the accuracy metrics which evaluates the forecasting performance from different perspectives.

2) Horizon of 4 also acts as a turning point for the ranking of three strategies in most cases of RMSE, PWE and Outbreak MAE. The iterated strategy usually ranks well for short-term prediction(H = 2 and 3), while MIMO strategy has advantages over longer prediction horizons (H $\geqslant$ 4).

3) Iterated strategy derives the least PWE over the majority of horizons with MLP as well, though it loses to MIMO strategy implemented with SVR.

The performance of different strategies demonstrated in the above results is in compliance with the theoretical characteristics of multi-step-ahead modelling strategies mentioned in previous literature[55,56]. Specifically, the iterated strategy accumulates errors of the trained single-output model at each step[55], which leads to relatively large prediction errors with the increase of forecasting horizon. However, if the ML model performs well for one-step ahead forecasting, this strategy can be quite useful for short-term multi-step-ahead prediction, like H = 2 and 3 in this context. Moreover, considering the mechanism of iterated strategy, it can capture some possible future information by adopting the predicted value as part of input variable, and thus could be more sensitive to the appearance of outbreaks and peaks, which may result in its superior performance considering PWE. While MIMO strategy preserves the stochastic dependency among the predicted sequence of values, which contributes to modeling the underlying dynamics of the time series[28,56], and thereby can be more suitable for longer prediction horizons.

## 6. Conclusions

Multi-step-ahead influenza prediction is of great significance on planning appropriate actions and timely responses to the potential influenza outbreaks and epidemics. However, prior researches seldom consider the different strategies for multi-step-ahead forecasting when implementing models for influenza prediction. Addressing this research gap, this study explores the performance of three basic multi-step-ahead modelling strategies (i.e. iterated strategy, direct strategy and MIMO strategy) implemented with machine learning models, using weekly ILI data of northern and southern China. In addition, to ensure the high quality of each model under each strategy, a CLPSO based modeling framework is developed to effectively and efficiently accomplish the model tuning tasks conducting parameter optimization and feature selection in a unified optimization process. In terms of performance evaluation, besides the commonly used statistical metrics such as MAPE and RMSE, the measurements aiming at the prediction performance during influenza outbreak seasons are also taken in order to fully analyze and assess the different strategies. Results of the experiments on two well-established machine learning methods, SVR and MLP, show that MIMO strategy demonstrates the best comprehensive forecasting performance over a wide range of prediction horizons while iterated strategy has advantages for decreasing the time error between the occurrence of predicted peak value and true peak value during an influenza outbreak period. Furthermore, different strategies are appropriate for the prediction horizons of different length. In the context of this study, several evaluation metrics indicate the similar trends that iterated strategy is superior for short-term influenza prediction over the horizons less than four, while MIMO strategy is more competitive for longer horizon.

We hope this study could result in more considerations of applying multi-step-ahead modelling strategies and provide a reliable reference for the selection of appropriate strategy when conducting multi-step-ahead influenza prediction, which could further contribute to the decision-making in the domain of infectious disease control and medical management. As for the limitation of this study, the conclusions are drawn based on univariate prediction model, merely using the lags of ILI rate as input variables. Since the main purpose of this study is to analysis the distinction of prediction performance under different multi-step-ahead strategies, the results could be more reliable without the interference from exogenous variables which may have complex relationship with the predicted target. However, for the purpose of appropriately choosing multi-step-ahead strategies, future efforts could be made to explore the improvement and development of hybrid modeling frameworks for multi-step-ahead influenza prediction involving commonly used exogenous variables such as weather condition, search engine query and social media data. Optimization algorithms for model tuning tasks in the framework could be refined for more complex situations as well.

## Acknowledgment

This work was supported by Natural Science Foundation of China under Project Nos. 71571080 and 71871101.